# User-Centered Feedback Design in Person-following Robots for Older Adults

Samuel Olatunji, Tal Oron-Gilad, Vardit Sarne-Fleischmann, Yael Edan

*Abstract*— **Feedback design is an important aspect in person-following robots for older adults. This paper presents a user-centered design approach to ensure the design is focused on users' needs and preferences. A sequence of user studies with a total of 35 older adults (aged 62 years and older) was conducted to explore their preferences regarding feedback parameters for a socially assistive person-following robot. The preferred level of robot transparency and the desired content for the feedback was first explored. This was followed by an assessment of the preferred mode and timing of feedback. The chosen feedback parameters were then implemented and evaluated in a final experiment to evaluate the effectiveness of the design. Results revealed that older adults preferred to receive only basic status information. They preferred voice feedback over tone, and at a continuous rate to keep them constantly aware of the state and actions of the robot. The outcome of the study is a further step towards feedback design guidelines that could improve interaction quality for person-following robots for older adults.**

## 1 INTRODUCTION

Socially assistive robots (SARs) are being developed to assist older adults in a wide range of activities. A major effort is focused on instrumental activities of daily living (IADLs), tasks that are not mandatory for fundamental functioning but essential for independent living and interaction with the environment (e.g., activities like housekeeping, or shopping) [1]. Some of these activities can be made easier for older adults with the assistance of a person-following robot. The robot can be programmed to autonomously track the older adult and follow as he or she moves while providing assistance. It often has a compartment to carry the belongings of the user as it follows. This relieves the older adults from the physical stress of carrying loads while walking and performing other IADLs [2]. The robot can also serve for the purpose of safety monitoring and companionship whilst supporting the older adult to maintain their independence in the home and outside.

Person-following is an important aspect in many service robotic applications [3] but it should be designed to conform with social norms and cultural values in order to inspire confidence and acceptability in the users. To create robots that move in socially acceptable manners, it is important to consider a multitude of parameters such as the robots' speed, acceleration and deceleration properties, the lead human's walking speed, and the appropriate physical proximity, as a function of the environment (e.g., a narrow corridor vs. an open room), context (e.g., routine vs. urgent), the user's physical state and intentions [4]–[6]. In addition to the robot's movement, there are other crucial components in the user's interaction with the robot that can affect the quality of the interaction (QoI) [7].

Identifying and addressing these crucial components require user studies to improve and ensure smoother human-robot interactions in person-following robots [8]. This is particularly critical for older adults who have peculiar needs that require attention [9], [10]. Some of these needs could be perception-related such as decline in visual, audial and haptic acuity [11]. Needs are also related to cognitive challenges that affect the rate of understanding, integrating and processing of information [12]. Physical challenges connected with stability and movement limitations also require special consideration during design [10]. SARs designed for older adults must therefore cater for their needs to ensure that the age-related peculiarities do not partially or completely limit their use.

## 2 RELATED WORK

Successful interaction requires communication between the human and the robot which generally involves sending and receiving of information to achieve specific goals [13]. Communicative actions when presented in the most comprehensible form promote understanding which aids a successful interaction of the user with the robot [14], [15]. The communicative actions from the robot to the user, herein referred to as feedback, are the presentation of information by the robot to the user in response to user's actions.

The content of the feedback information provided is an essential influencing factor for successful interaction between humans and robots [16]. Feedback content is predicated on the desired level of transparency (LOT) [17], [18]. LOT, in this context, can be described as the degree of task, environment, robot, human, and interaction-related, information provided to users while the robot is performing its task [19].

Task-related information consists of information provided by the robot to inform the user of its state, or its actions in relation to the task. It also includes information on the reasons for actions taken while executing the task, the next actions to be taken and the progress of the task. This was demonstrated in the situation awareness based transparency model (SAT) for autonomous systems developed by Chen et al., [20] which mirrors Endsley's model of situation awareness [21]. An adaptation of that model in relation to a person following robot is presented in *Table 1.*

Table 1: Information provided at various LOT

| LOT | Information Provided |
|---|---|
| 1. Perception | Information about **the state of the robot** and/or contextual information that the user must be aware of. For example – the robot makes a sound or says 'yes' when it acknowledges the user giving the command 'follow me'. |
| 2. Comprehension | Information about **how** the state of the robot or the context may affect achieving the goal For example – The robot verbally says that it is following the user from behind in a distance of 2 meters. |
| 3. Projection | Information about **how the future state** of the robot may change based on the context For example – The robot verbally says that in a few meters it will have to slow down to an anticipated change in the walking surface. |

Environment-related information which the robot could provide include constraints of the environment, type of environment and any other safety-related information about the environment [6], [19], [22]. Robot-related information includes information from the robot regarding its degree of reliability, underlying principles of its decision making and all other information pertaining to the robot (for example, information on battery status, operating mode, how to use a specific feature on the robot etc) [23].

Human-related information includes the human's physical and emotional state if the robot can assess it. It also includes information regarding the human's effort in the task, workload or stress encountered if it can be provided by the robot [24]. Interaction-related information involves details of the roles of the robot and human in the interaction, shared awareness and dynamics of the teamwork [24], [25].

Implications of providing information to users was explored in [26]. They suggested that a robot which is truly transparent may contravene the ideology of worthy companionship where the companion has a social value of independence, agency and autonomy to disclose information. The authors hypothesized that as transparency is increased the user may perceive the robot more as a tool than a companion. This is contrary to the expectation desired in domestic and healthcare settings where the users are expected to interact with the robots as partners, companions and entities capable of caring for them. It was recommended that various levels of transparency in the robot's communication should be evaluated in a wide range of domestic environments to explore the relationship between transparency, utility and trust for HRI [26].

How the robot communicates is also a crucial component of the interaction in relation to what information is being communicated [26]–[28]. The information can be presented in various modes such as audial, visual or haptic modes [12], [29]. It could also be in various other forms of non-verbal modes such as eye blinks, shifts in gaze (for robots with a face) or body posture for humanoid robots [30]. Implicit non-verbal communication positively impacts understandability, efficiency and robustness to errors arising from miscommunication [31]. Transparency often helps to reduce the conflict in joint task situations when such errors occur [31]. The effect of transparency and communication modality on trust was examined in [32]. The modality was not significant in this study which included a simulated robot deployed on a desktop computer, though the transparency manipulation was significant. This interaction differs from interaction with a mobile and embodied robot such as a person following robot which this current paper focuses on. Also, the users in [33] were undergraduate students (aged 18 – 22 years) which have different characteristics and perceptual peculiarities from the older adults. It was recommended that more user studies should be carried out in specific domains in order to determine influence of information level, modality and content on trust.

When discussing strategies to foster transparency between the human and the robot, it was recommended that the interface through which the human interacted with the robot should provide useful information relating to the task and environment [19]. The author cautioned that too much information or a non-intuitive display may cause confusion or frustration for the user [19]. This is in agreement with the findings in [34] where it was additionally noted that multi-modal communication aided performance of the users. Though Kim and Hinds [35] remarked in their study that users understand the robot better if it explains the reasons behind it's behaviour. This was confirmed by [36] in an unmanned aerial system scenario with multiple operators. The hypothesis should be further investigated in other scenarios to determine if this varies with the complexity and nature of the task or environment. Studies conducted in [19] added that cues to signify what the robot was doing, its reliability status and the presence of a face on the robot, help the user trust the robot better. It was noted that style and modality of communication with the inclusion of some etiquettes, robotic emotional expression and gestures in the feedback could influence performance of users using the automated system [19].

Timing of the feedback is also critical to maintain comprehension of the information being communicated [37]. For instance, feedback given too late causes confusion [18]. Temporal immediacy between a user's input and the robot's response influences the naturalness of the interaction [38]. To increase the trust of the user it is important to provide continuous feedback regarding the reliability of the robot [39]. This agrees with the findings in [33] regarding providing continuous stream of information. The question of which information to provide continuously and which information to reserve for the user's demand arises. This often varies based on the type of task, feedback modality, and the category of potential users as observed in [40]. The preference of the user regarding feedback timing along with the content and mode of feedback in specific tasks is essential to foster smoother coordination and collaboration between the human and the robot.

In previous studies involving person-following robot applications, most of the developments did not explicitly incorporate feedback from the robot regarding the robot's actions as it follows. The robot simply followed the target person as soon as the person was detected in a predetermined range as noted in [41]. This caused confusion for many of the participants regarding what the robot was doing per time. Several participants were unsure if the robot was following them, stopping or had lost track. This lack of communication from the robot could lead to a loss of SA which makes the users uncertain or unsure of the state of the interaction at each point in time [42]. This has the potential to degrade the interaction quality of the older people with the robot. The few studies that incorporated feedback [43]–[45] provided message acknowledging user commands such as saying 'yes' or other specific expressions [44]. These were implemented as part of the robot's behaviour without explicit user studies to determine the preferred content, mode or timing of feedback from the robot. There is generally a gap in user-centered preferences in design of feedback for person following robots [5], particularly those used in eldercare [12].

The current study presents a user-centered design approach to ensure the design is focused on older adults' needs and preferences. Older adults' preferences for feedback design were evaluated in a series of empirical studies. Feedback design was constructed consecutively, looking first at the preferred robot LOT (perception, comprehension, projection), and the content of the information to be presented (depending on the LOT), followed by the mode (voice or tone) timing and frequency of the feedback (continuous or discrete). It is crucial to note that while identifying preferred feedback parameters, individual differences come into play [46]. There are several sources of individual differences in older adults which usually have potential implications in the design process [11], [46]. Two of the sources, which were considered in this study were age and gender due to the fairly balanced distribution of these factors in the older adult population recruited. The influence of these factors in the feedback design was highlighted through the analyses. The aim was to improve quality of interaction taking into account the different age groups and gender, while ensuring increased user satisfaction and acceptance. This paper presents comprehensive analysis of our previous study [47] which highlighted the importance of the feedback design considerations but did not provide the details. Additionally, we provide new analyses regarding the influence of gender and age on the feedback design. Finally, the paper presents design guidelines for feedback design in the development of an assistive person-following robot for older adults.

# 3 METHODS

## 3.1 Overview

The study was constructed with a coactive design perspective. It involved preliminary discussions with older people on their expectations about a robot in this context of person-following. This aligned the thoughts of the potential users and designers into the same conceptual design zone ensuring robot performance is tuned to meet the users' expectations. The highlights of these preliminary discussions bordered on the overarching goals of the robot such as what the robot does, why and how. This laid the foundation of the intentional model of transparency on which the specific task related model of transparency addressed in the current study was built. Preliminary experiments were also conducted to explore proxemics and movement preferences of the robot as it follows the user in different environmental conditions [6]. These provided some environment related preferences, constraints and context which guided the feedback design options in the current work. In this research, a sequence of experimental user studies with older adults were performed to evaluate step by step:

- *What level of transparency would the older adults desire and what would they prefer as feedback content at their desired LOT?*
- *Which feedback mode would the older adults prefer?*
- *What would the preferred feedback timing be?*

The design parameters gathered in these user studies were implemented and tested in a final experiment to evaluate the effectiveness of the feedback design. This experiment evaluated if *the feedback implementation improves the quality of interaction.*

## 3.2 Apparatus

A Pioneer LX mobile robot (50 cm width, 70 cm length and 45 cm height) equipped with an integrated on-board computer, 1.8 GHz Dual Core processor, and 2GB DDR3 RAM was used.

A built-in SICK S300 scanning laser rangefinder, mounted approximately 20 cm above the ground, was used to detect nearby obstacles and stop the robot if it detected an object 50 cm from its core. Distant obstacles were tracked using an external Kinect camera with a pan mechanism that was added to the robot and mounted 1.5m from the ground, as shown in *Figure 1*. The person tracking and following commands were developed and executed in ROS [47] and were sent to the Pioneer LX's onboard computer using a TPLINK router with wireless speed up to 300 Mbps.

## 3.3 Algorithm Development

The algorithm works without a map. This is important to ensure flexible operation in a multitude of environments. The map of the environment used in this study is presented in *Figure 2*. Open Track [48], [49] is used to identify and track the coordinates of the person to be followed. Some adjustments were incorporated to ensure it can detect a human 1.4m to 2m tall, with a confidence level threshold of 1.1. The robot selects the first person detected and moves the robot to the defined position behind the person. Continuous estimation of the person's position is achieved using the robot's pan angle ($RAngle$) and the angle of the detected person ($PAngle$), measured from the centre of the robot, to constantly estimate the position of the person. The person's position (coordinates X,Y) is calculated as follows:

$$X = distance\big(cos(Pangle + RAngle)\big) \tag{1}$$

$$Y = distance\big(sin(Pangle + RAngle)\big) \tag{2}$$

The linear velocity ($lVel$) of the robot is updated dynamically based on the distance between the robot and the target while the angular velocity ($aVel$) is updated dynamically based on the angular displacement of the target.

These are calculated as follows:

$$(lVel) = (lVel)(distance - target) \tag{3}$$

$$(aVel) = (PAngle)(aVel) \tag{4}$$

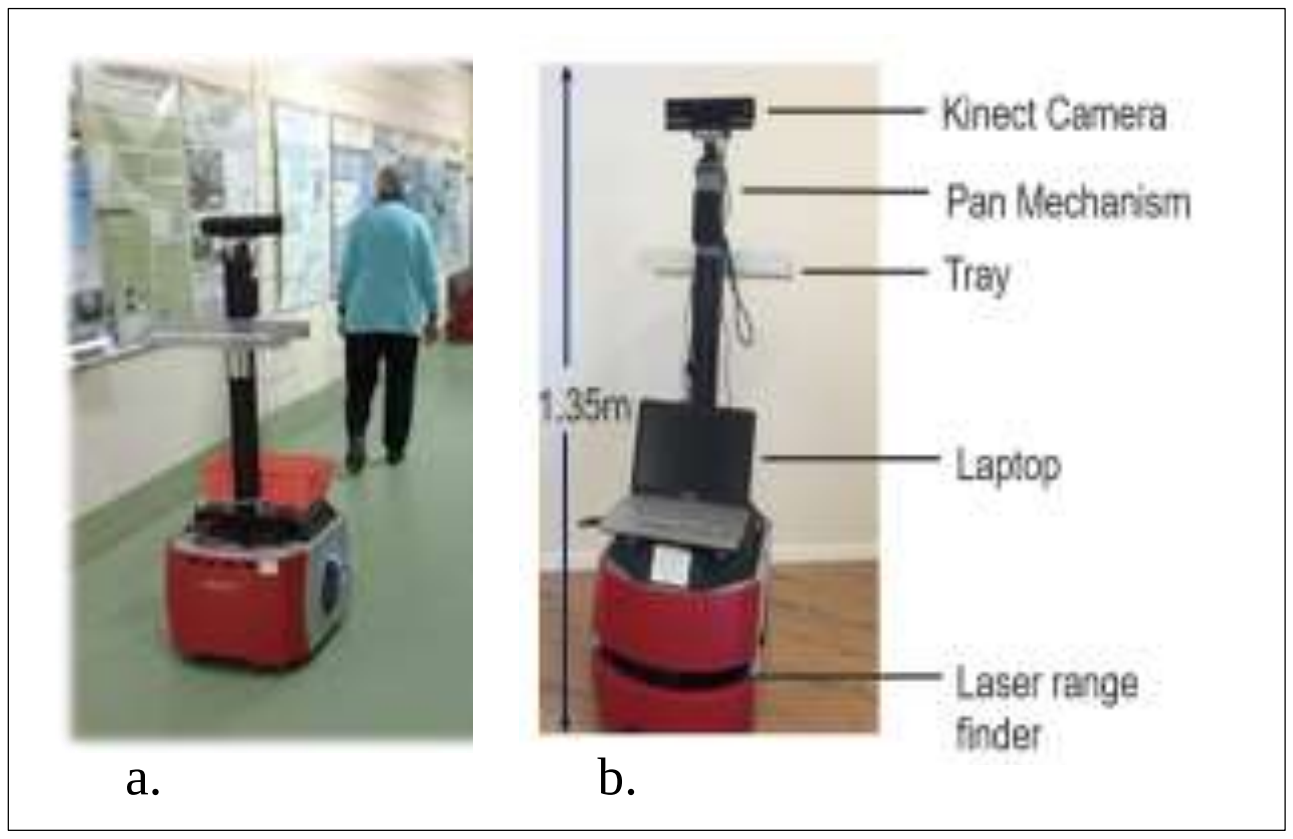

Figure 1: (a) A participant being followed by the robot (b) Experimental platform – Pioneer LX mobile robot.

Parameters were set according to recommendations for social following robots [3], [5], [21], [26], [51]. The maximum following speed was set to 1.0m/s for safety reasons as emphasized in [6], [52], [53]. Other parameters such as acceleration coefficient, following distance and following angle were set to 0.5, 0.3m and 30˚, respectively. A summary of the algorithm is presented in Table *2* below.

Table 2: Person detection and following algorithm pseudocode

**Algorithm for the Person detection and following**

**Initialize:**
$\boldsymbol{linear_velocity}\ (l_{vel}), \boldsymbol{angular_velocity}\ (a_{vel}),$
$\textbf{following_distance}\ (\mathbf{d}), \textbf{following_angle}\ (\mathbf{a}),$
$\textbf{acceleration_coefficient}\ (\mathbf{a})$
**for** each person detected [ **i** ], **do**
estimate position of person: $x_{person}, y_{person}$
compute angle error of robot:
$$y_{error} = y_{person} - sin\ (a)\ d$$
$$x_{error} = x_{person} + 0.4 - cos(a)d$$
$$angle_{error} = tan^{-1}(\frac{y_{error}}{x_{error}})$$
compute distance error of robot:
$$dist_{error} = \sqrt{x_{error}^2 + y_{error}^2}$$
update vel. of robot with prop controller:
$$l_{vel} \leftarrow dist_{error}(dist_{pcontrol})$$
$$a_{vel} \leftarrow angle_{error}(angle_{pcontrol})$$
safety measures; obstacle avoidance
send vel. update to robot motion planner
**end for** loop

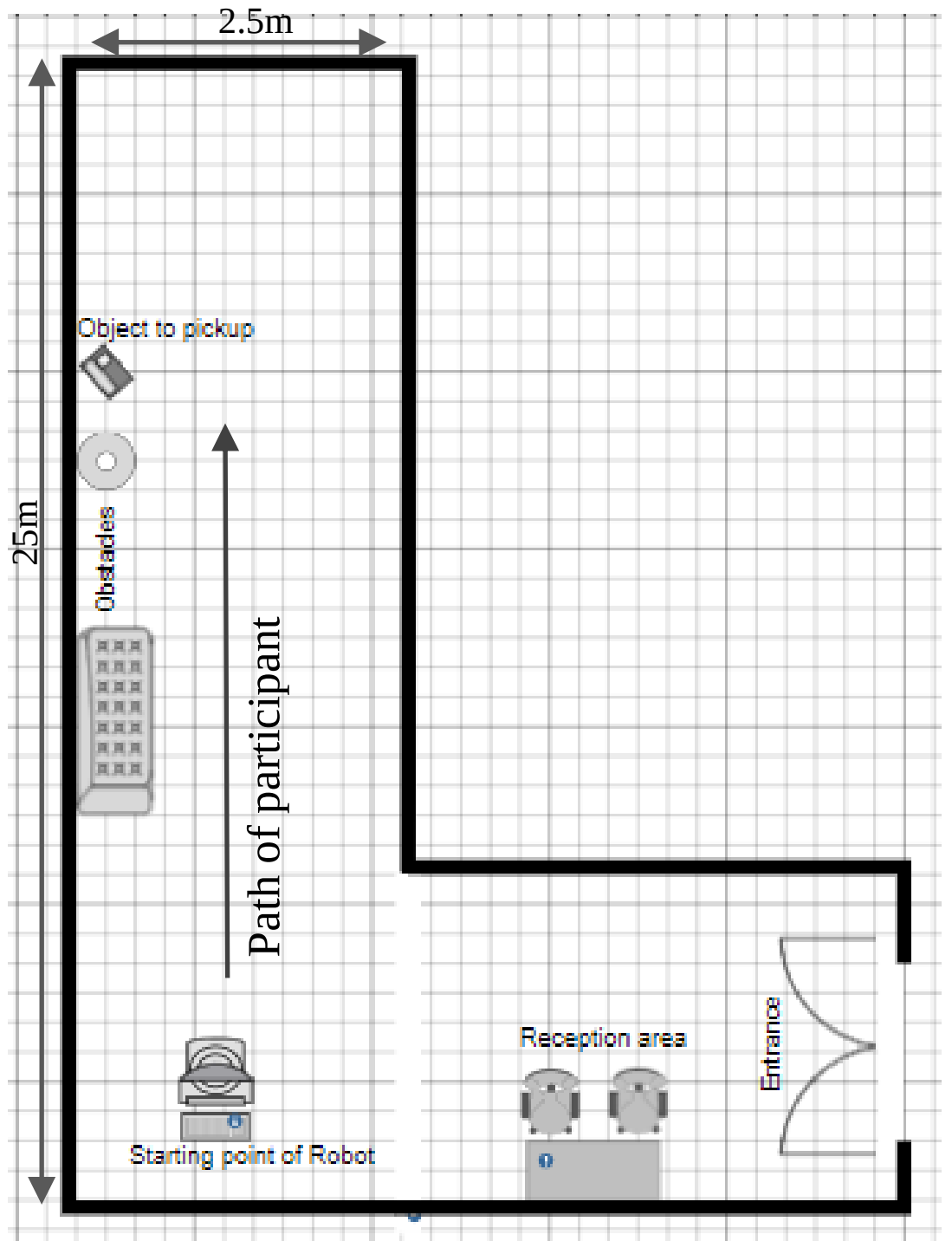

Figure 2: Map of the environment in which experiments were performed

## 3.4 User studies

A sequence of user studies was conducted as presented in *Figure 3*. Each experiment was independent with the outcome preferences from each experiment implemented in the consecutive experiment.

## 3.5 Procedure

Before each experimental session, participants completed a preliminary questionnaire. This included demographic information, the Technology Adoption Propensity (TAP) index [54] and the Negative Attitude toward Robots Scale (NARS) [55]. They were then introduced to the robot and to the experimental task. The task was to walk down a straight 40m path while the robot communicated with them by voice in English. The audio feedback was provided directly by the robot's speakers which produced a sound of approximately 60dB above the noise level in the building which was about 35dB. The robot followed at a specified distance, angle and speed (section 3.4). The study took place in a 2.5 m wide corridor in a university laboratory building.

A trial refers to each session when the participants interacted with the robot which includes walking the designated path with the robot to retrieve a specified item at a specified location. After each trial, a condensed form of Situation Awareness Rating Tool (SART) [56] was used to assess the level of situational awareness and understanding the participants had in each session. This was administered along with some other questions relating to the the preference of the participants in each session as used in [29]. The post-trial questionnaire used 3-point Likert scales with 3 representing "Agree" and 1 representing "Disagree". The 3-point scale was selected based on previous trials with older adults that revealed that they experienced difficulty and sometimes confusion representing their opinion on the 5 and 7 point scales [57]. At the end of all trials, a final questionnaire was provided to enable the participants to express their opinion regarding the experience with the robot. All procedures were approved by the university's ethical committee.

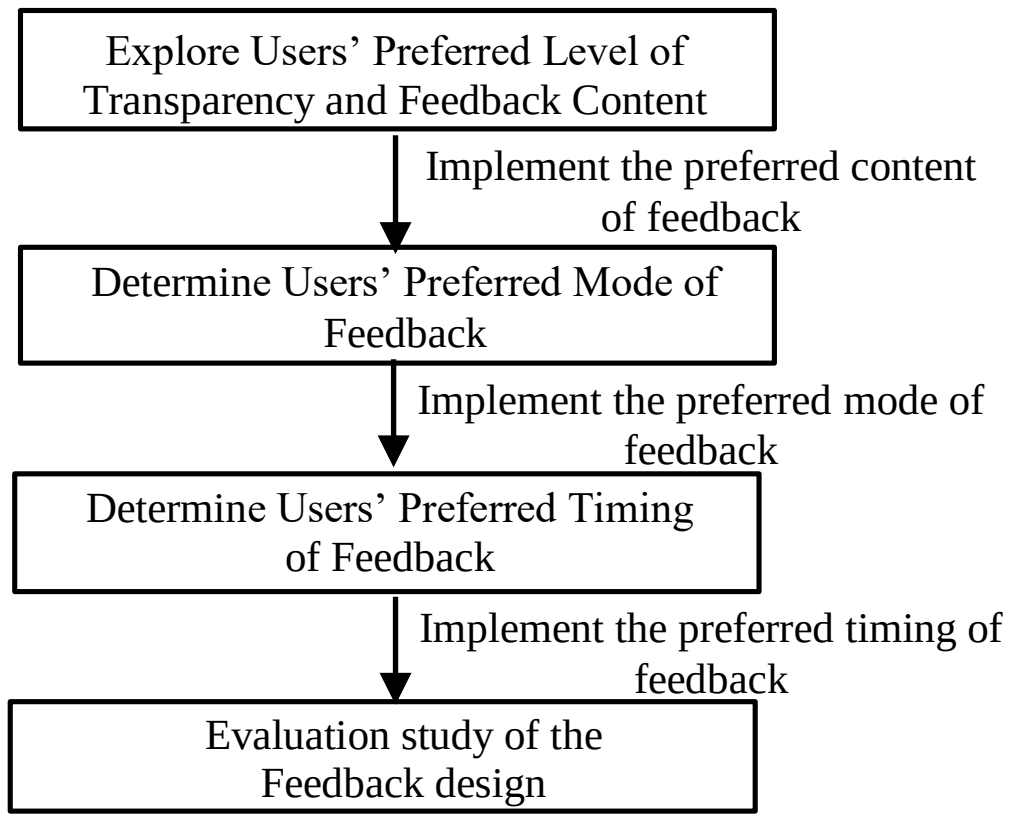

Figure 3: Experimental flow of the User Studies.

## 3.6 Analyses

The analyses were performed using the following objective and subjective measures acquired during the experiments as detailed below:

*Objective measures*: Understanding was measured as the number of clarifications participants made by participants which caused interruptions. An interruption in this context is the period during the experiment when the participant does not understand the information the robot is presenting or what the robot is doing and therefore pauses to ask the experimenter questions for clarification regarding the information the robot was giving. Another measure of understanding was the reaction time which was measured as the time it took participants to respond to the robot's instructions. Effort was measured based on participant's heartrate before and after each trial. The heartrate was measured with a Garmin watch Forerunner 235 series. The watch was worn by the participants at the start of the experiment and till the end of the experiment. The heart rate readings (bpm) was taken at the start and end of each trial. Then, the heartrate variability was measured and used in the analyses. Engagement was measured based on the duration of gazes the participants made to the robot during communication, total time spent with the robot, and number of times participants initiated communication with the robot while gazing at the robot), Trust was measured via the overall time spent on the task of walking to pick item without looking back at the robot coming behind and the time spent waiting for the robot when the robot lost track or delayed. Comfortability was measured as the number of times participants were glancing back at the robot.

S*ubjective measures:* Users' responses regarding their level of understanding, comfort, engagement, persuasiveness and satisfaction were assessed through questionnaires and short interviews at the end of each experimental trial.

*Data Analyses*: The tests were designed as two-tailed with a significance level of 0.05. The model for the analyses was the General Linear Mixed Model (GLMM) with user ID included as a random effect to account for individual differences among participants.

# 4 Level of Transparency and Content of Feedback

The aim of this experiment was to provide the users sufficient situation awareness without overwhelming them with excess information. The preferable level of transparency was explored along with the appropriate information content.

## 4.1 Experimental design

*Independent Variables:* Level of transparency was the independent variable. Three levels of information were presented to the participant: **what** the robot is doing (e.g. 'following', 'stopping'), **why** the robot was doing what it was doing (e.g. 'stopping because the participant stopped, stopping because of an obstacle), and what the robot was **planning to do next** (e.g. informing the participants that it will stop whenever they stop).

*Dependent Variables*: Preference regarding the amount of information participants wanted the robot to present to them was collected through questionnaires and short interviews that contained specific items related to the participants' understanding of the robot's feedback, level of comfort and mental workload while interacting at various levels of transparency. The mental workload assessment was included due to the mental effort that could be required by the older adults to process the information the robot is presenting to them [12], [46]. The robot presented some information to the participants as described earlier. The participants were then asked for their preferences through questionnaires. They were also given the opportunity to add other expressions or information they would want the robot to give in addition to what it was already presenting to them. These responses from the participants were collected through questionnaires and interviews.

*Participants:* Thirteen older adult participants (8 Females, 5 Males) aged 65-85 were recruited via social networks and colleagues. They were all healthy participants with no physical disability, vision or hearing impairment. A short interview was held with them before the experiment commenced to ascertain their comfort with the experiments and understanding of the procedure. Each participant experienced all three levels of information presentation from the robot in random order. They completed the study separately at different timeslots, so there was no contact between participants.

## 4.2 Results

Analysis on LOT preferences (*Figure 4*) revealed significant differences among users ($M=1.62$, $SD=0.87$, $p<0.001$). All of the participants preferred the robot to say what it was doing at the moment (LOT level 1). 38% ($M=0.38$, $SD=0.5$) of the participants wanted the robot to additionally present the reason for its actions (LOT level 2), while only 23% ($M=0.23$, $SD=0.44$) of the participants wanted information on future actions of the robot (LOT level 3).

Participants did not express discomfort or excess workload while interacting with the robot at higher LOTs. They also gave their preferences for specific feedback content from the robot (*Figure 5*). Several participants wanted it to say more than basic task related information such as 'following' or 'stopping'. Some wished it would introduce itself and greet them. Most of the participants (85%) also desired for the robot to communicate in their native language (Hebrew).

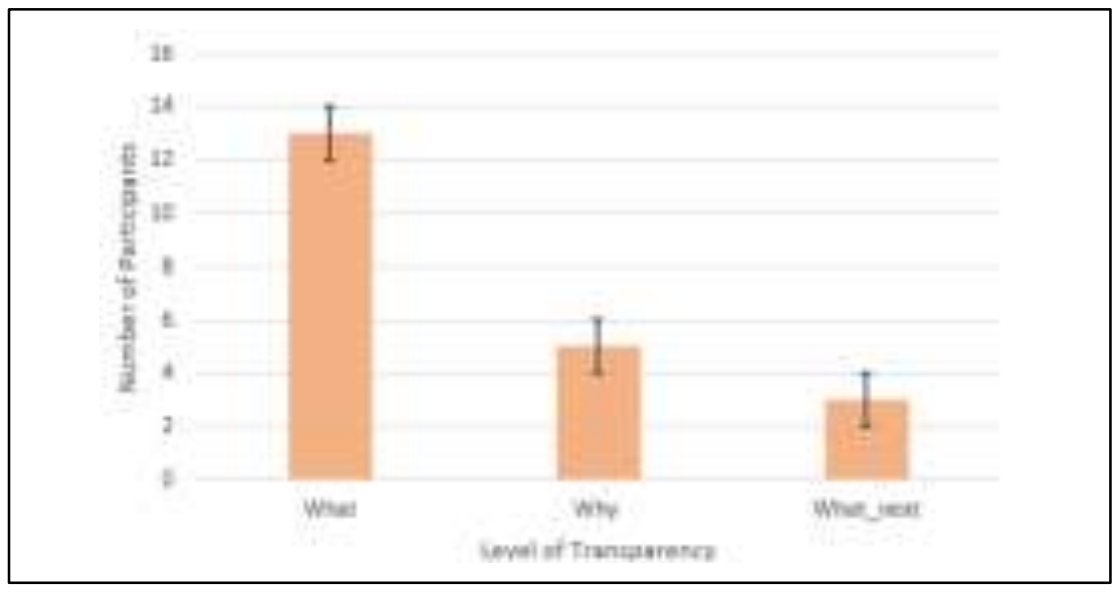

Figure 4: User preference regarding level of transparency

The results provided the rationale for the use of the first LOT (robot's current action) with specific expressions such as 'starting', 'following', 'stopping' in the next experimental stage. Greetings according to the suggested content (such as 'Hello' or 'Bye') during the interaction with the robot were added to the communication to make it friendlier. This modification was implemented for subsequent studies by enabling the users to choose the preferred language of feedback (English or Hebrew).

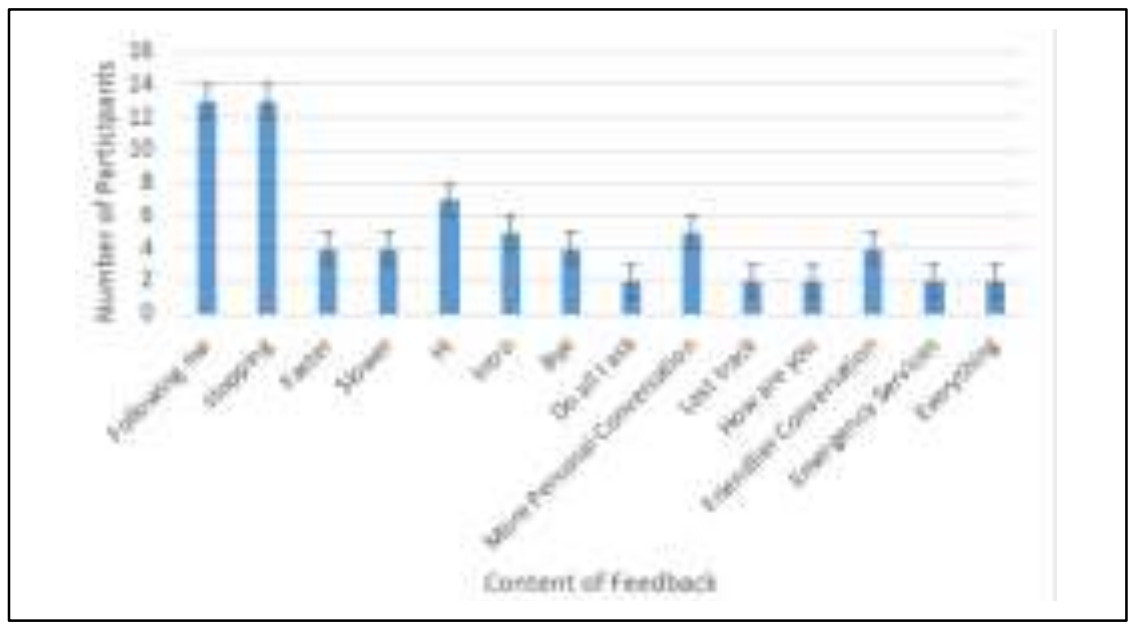

Figure 5: Users' suggestions for content of feedback

## 5 MODE OF FEEDBACK

The aim of this experiment was to identify the most suitable mode of feedback considering that the robot is specifically a person-following robot which would be behind the user most of the time. This requires the feedback to be audible to the user particularly when following. Two audial feedback modes were explored: a female voice (as recommended in [11] and [12]) and a tone in form of a sequence of beeps (beep, beep…). The voice content was: 'following', 'stopping', and greetings. The voice was in the form of a recorded human speech in order to obtain as sound as close as possible to natural human communication. The beeping started once the robot began to follow and ceased when it stopped. The sound of the voice and tone feedback was maintained at approximately 60dB, well above background noise level. The volume was made adjustable to the preference of the participant, such that it could be increased or decreased to make it comfortable and audible to the participant in accordance with audial feedback design guidelines provided [58].

The feedback modes were implemented according to design guidelines for general multimodal human-robot interaction [59]. The standards for developers to address the needs of older persons [11], [12] was also consulted in order to satisfy design recommendations for presentation of auditory information. Actual human speech was used instead of synthesized speech based on earlier studies which revealed that it aided higher intelligibility [60]. A native speaker's recording was used in order to avoid accent-related understanding difficulties [61]. The content of the feedback was based on the results obtained in the previous stage.

### 5.1 Experimental Design

*Independent Variables:* The mode of feedback manipulated as voice mode and tone mode.

*Dependent Variables:* Subjective and objective measures as described in section 3.7.

*Participants:* Twelve additional older adults' participants (9 Females, 3 Males) aged 62-73, were recruited. They were physically and cognitively fit for the experiments as described in section 4.1. Each participant received feedback from the robot in both tone and voice modes.

### 5.2 Results

Analysis revealed that 10 of the participants (77%) preferred the voice feedback mode (*M=0.77*, *SD=0.43*) to the tone mode (*M=0.08*, *SD=0.272*) and 8% were fine with either of the modes (*M=0.15*, *SD=0.368*). This effect of feedback mode on their preference was significant (*M=0.92*, *SD=0.484*, $p<0.001$). Feedback mode had no significant effect on comfort, engagement and persuasiveness. Eight of the 12 participants reported that they were comfortable in both trials. Three of the participants were indifferent. This outcome is illustrated in *Figure 6*.

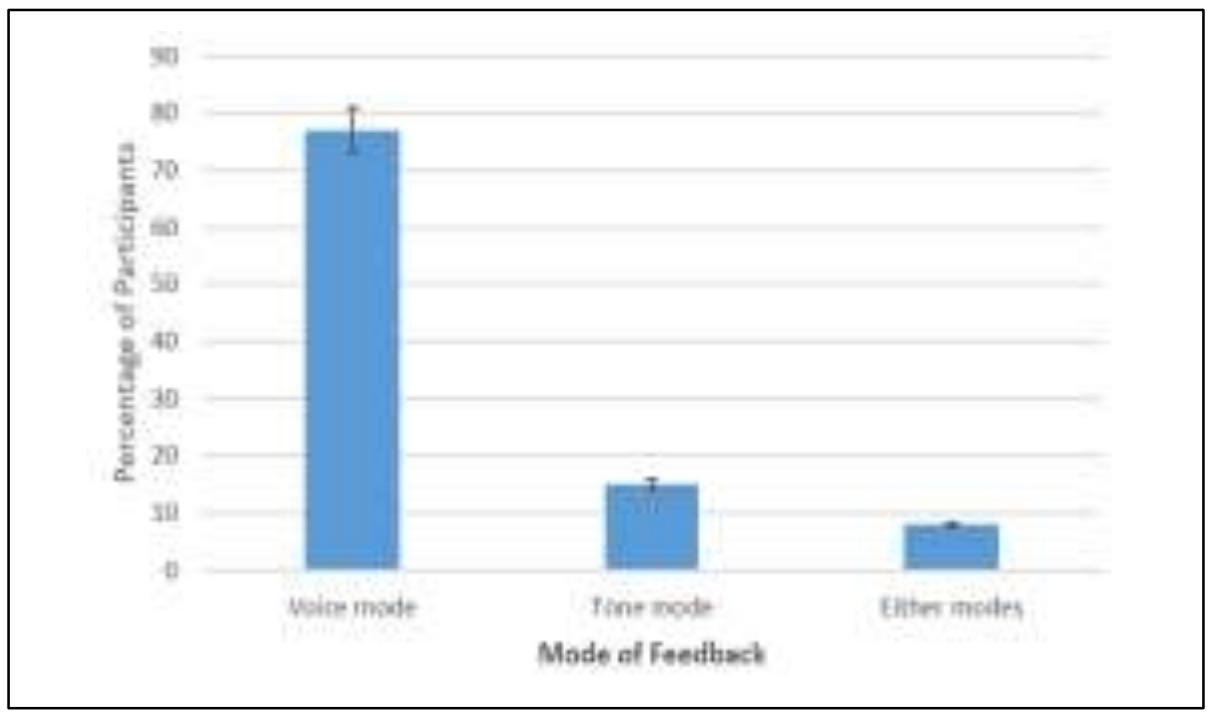

Figure 6: Users' preferred mode of feedback

The heart rate variability was also not significantly affected by the feedback mode. A one-way ANOVA using mode of feedback as the fixed factor and user ID as a random effect revealed that the mode of feedback had a statistically significant effect on the users' **understanding** (*M=2.0*, *SD=0.938*, $p<0.001$). Voice feedback was therefore used for the subsequent experimental stages.

## 6 TIMING OF FEEDBACK

The temporal dimension of the feedback preference of the older adults was studied. The transparency level, content and mode of feedback were based on the outcome of the previous stages.

### 6.1 Experimental Design

*Independent Variables:* the timing of feedback included three timing options: continuous (5 and 10 seconds intervals) and discrete. As an example, in the continuous timing mode (5 seconds interval), the verbal feedback was given continuously, every 5 seconds (e.g., "Following", "Following", every 5 seconds). In the discrete timing mode, the feedback was given only at the beginning and at the end of the interaction with the robot. In this mode, the robot would simply inform the participants when it begins the following and inform the participants when it is stopping.

*Dependent Variables:* the same variables described in section 3.7.

*Participants:* The same 12 participants recruited in 5.1 followed up in this experiment. Each participant received verbal feedback from the robot in the discrete and continuous timing options. They answered brief questions in questionnaire and interview format after the trials regarding which feedback timing they prefer and why.

### 6.2 Results

Analyses (**Error! Reference source not found.**) showed that 80% (10) of the participants preferred the continuous feedback (*M=0.85, SD=0.366*) over the discrete feedback with (*M=0.15, SD=0.366*). The effect of the feedback timing on the users' preference was statistically significant *(M=1.46, SD=0.756, p<0.001*). The effect of feedback timing as a fixed variable on understanding was also statistically significant (*M=1.87, SD=0.923, p<0.001*). Among those who selected the continuous feedback as their preferred timing mode, 84.6% preferred an interval of 5 seconds (*M=0.69, SD=0.468*) over 10 seconds (*M=0.15, SD=0.366*). The reason given was better awareness of what the robot was doing behind them at every point in time. This provided a rationale for the use of continuous feedback at the rate of 5 seconds in the following quality of interaction evaluation experiment.

## 7 DOES THE FEEDBACK IMPLEMENTATION IMPROVE THE QUALITY OF INTERACTION?

The feedback design parameters obtained in the three previous user studies were evaluated in light of their effects on the quality of interaction relative to a person following robot with no feedback.

*Hypotheses*: the feedback design implementation will improve the quality of interaction with the specific hypotheses stated as follows assuming that the feedback:
*H1: will improve the engagement of the participants.*
*H2: will increase the understanding of the robot for the participants.*
*H3: will improve the trust the user has in the robot.*
*H4: will improve the comfortability of the participant*.

### 7.1 Experimental Design

*Independent Variables:* There were two groups: one group interacted with the robot without feedback, the other group interacted with the implemented feedback.

*Dependent Variables:* Quality of interaction was measured both objectively and subjectively in terms of engagement, understanding, trust and comfort [7].

*Objective Measures*: Engagement and comfortability were measured as explained in section 3.7. The number of clarifications made by the participant during the interaction was counted as a measure of the understanding they had regarding the information the robot was giving them. A summary of the objective measures used are presented in Table *3*

Table 3: Objective Measures used for the Quality of Interaction

| *Variable* | *Objective Measures* |
|---|---|
| Engagement | Gaze duration (seconds) |
| Understanding | Number of interruptions to ask for more clarification (counted)<br><br>Reaction time (seconds) |
| Trust | Overall time spent on task of walking to pick item without looking back at the robot coming behind (seconds).<br><br>Time spent waiting for the robot when the robot lost track or delayed (seconds) |
| Comfortability | Number of times participants were glancing back at the robot (counted) |

*Subjective Measures:* Questionnaires and short interviews regarding their comfort level, understanding of the robot's information, trust and satisfaction as explained in section 3.6.

*Participants:* 20 older adult participants (13 Females, 7 Males) aged 65-85. They were healthy participants with no major physical disability. Ten of the participants received feedback from the robot while the other 10 received no feedback from the robot. Additional analyses were conducted to explore the potential influence the age and gender of the participants could have on the QoI variables assessed. The influence of age of the participants was assessed by conducting a correlation analysis between the age of the participants (*M=78.84, SD=6.72*) and the different QoI variables.

*Feedback Design*: Feedback was designed using the preferred parameters identified in the preceding stages as detailed in Table *4*.

Table 4: Parameters for Feedback Design

| *Parameter* | *Preference* | *Description* |
|---|---|---|
| Level of Transparency | Level 1 LOT | Information on what the robot is currently doing. |
| Content of Feedback | Action of the robot, Friendly content. | Specific information such as 'Starting', 'Following', 'Stopping'. Greetings from the robot. |
| Mode of Feedback | Voice Feedback | Audible female voice with speech rate less than 140 wpm with adequate pauses at grammatical boundaries. |
| Timing of Feedback | Continuous Feedback (5 seconds interval) | Notification of the state of the robot every 5 seconds (like, 'following, following..') |

### 7.2 Results

**Attitude Towards Technology**

Most of the participants were acquainted with the use of innovative technologies (*M = 3.39, SD = 0.72*). The TAP index [20] revealed that more than half of the participants were affirmative that technology could provide more control and flexibility in life (*M = 2.48, SD = 1.59*). Several of them also showed confidence in learning new technologies (*M = 2.95,*

*SD = 1.18*), and trusted technology (*M = 3.04, SD = 1.58*). The NARS index [21] revealed that about 60% of the participants felt that if they depended too much on the robot, something bad might happen (*M= 3.05, SD = 1.19*).

**Engagement**

The results revealed an increase in the time (*M=3.15, SD=4.38)* the participant was focused on the robot while the robot was presenting some information about the interaction before following (*F(1,37) =20, p<0.001*). In the group where the feedback was implemented, it was observed that participants were willing to spend more time communicating with the robot (*M=5.65, SD=4.65*) compared to the group without feedback (*M=0.53, SD=1.88*). There was a 60% improvement in the communication frequency suggesting improved engagement.

Responses from the questionnaire also showed significant differences in the response of the participants related to engagement (*M=2.38, SD=0.74, p<0.001*). Participants in the group with feedback (*M=2.76, SD=0.54*) made more positive comments regarding the naturalness of the robot that made them feel more connected to the robot compared to those in the group without feedback (*M=1.96, SD=0.72*). Several of the participants in the group with feedback expressed excitement at the robot's communicative ability. Some of the comments made were: "I was thrilled to hear the robot communicate with me in Hebrew. It helped me relate better with it", "the way it spoke every time, telling me what it's doing made it interesting to interact with". These comments suggest some form of engagement with the robot.

**Understanding**

The understanding of the participants improved with the feedback design as expressed by the amount of time (*M=2.84, SD=3.81)* the participants impeded the flow of the interaction due to clarifications they were making regarding the actions of the robot (*F(1,37) =3.7, p<0.062*). Participants in the group with feedback experienced a smoother flow in the interaction with minimal interruptions (*M=0.75, SD=1.12*). Participants in the group without feedback interrupted the flow of the interaction more frequently when they were not certain of what the robot was doing (*M=5.05, SD=4.39*).

In terms of the reaction time (*M=2.2, SD=1.84, p=0.013*), participants in the group with feedback (*M=2.9, SD=1.94*) spent more time (seconds) listening to the instructions from the robot before taking action (*F(1,37)=6.76, p<0.013*) compared to the group without the feedback design (*M=1.47, SD=1.42*). Additionally, the participants' responses in the questionnaires regarding understanding (*M=2.32, SD=0.66, p<0.001*) showed that the group with feedback (*M=2.76, SD=0.54*) had a better understanding of the robot than the group without the feedback (*M=1.83, SD=0.37*).

**Trust**

Results revealed that the participants focused on the task without worrying about the robot coming from behind when the feedback was implemented (*M=76.95, SD=20.08*), as seen in the time they spent in the task of picking up an item (*M=99.33, SD=32.95).* This was statistically significant, (*F(1,37)=36.78, p<0.001*) compared to the time spent in the group without feedback (*M=122.89, SD=26.9*). This suggests they gained some level of trust that the robot would not collide with them or cause any harm to them.

Participants in the group with feedback (*M=0.75, SD=1.12*) waited (*M=2.85, SD=3.81*) less compared to participants in the group without feedback (*M=5.05, SD=4.4*). This could have been due to better awareness of what the state of the robot was if it was delayed or lost track. This was statistically significant (*F (1,37) =18, p<0.001*).

**Comfortability**

Regarding comfortability as measured by the number of back glances (*M=2.9, SD=3.6*) the participants took, there was no significant difference (*F(1,37) =0.073, p=0.88*) between the participants in the group with feedback (*M=3, SD=4.05*) and those in the group without feedback (*M=2.68, SD=3.15*). A significant difference was also found in the comfortability of communicating with the robot (*M=2.37, SD=0.7, p=0.004*). Participants in the group with feedback (*M=2.67, SD=0.66*) responded more positively regarding comfortability with the robot than those in the group without feedback (*M=2.05, SD=0.66*).

The influence of the feedback on the QoI variables as measured in the objective variables is presented inTable *5*.

Table 5: Significant Influence of feedback implemented on QoI variables

| | | ***Engage*** | ***Understand*** | ***Trust*** | ***Comfort*** |
|---|---|---|---|---|---|
| NFD | Mean | *0.53* | *1.47* | *122.89* | 2.68 |
| | SD | *1.88* | *1.42* | *26.9* | 3.15 |
| WFD | Mean | *5.65* | *2.9* | *76.95* | 3.00 |
| | SD | *4.65* | *1.94* | *20.08* | 4.05 |
| | Sig. | *<.01*** | *.01** | *<.01*** | *.88* |

**p<0.05, **p<0.01, NFD = No Feedback, WFD = with Feedback, Engage=Engagement, Understand = Understanding, Comfort = Comfortability*

**Influence of Initial Attitude of Participants**

Correlation analyses were conducted to explore the possible relationships between the predisposition of the participants in form of NARS index and the objective variables. These were conducted using Pearson's Correlation Coefficient analyses to determine the trend, significance and effect size. A significant positive correlation was observed between the NARS index of the participants and engagement (*r=0.51, n=20, p=0.021*). Participants who had more negative reaction towards the robot gazed more intently at the robot.

There was also significant positive correlation between the NARS index of participants and the level of understanding in terms of number of interruptions made to ask for clarification (*r=0.491, n=20, p=0.028*) and reaction time (*r=0.448, n=20, p=0.047*). Participants who were more negatively disposed to the robot seemed to ask more questions about the robot and also had a longer reaction time to the robot's requests.

There was a negative correlation between the NARS index of participants and the trust the participants had in the robots.

The more negative predisposition the participants had regarding the robot, the less they trusted the robot as observed in the duration of time they spent on the task with the robot (*r=-0.558, n=20, p=0.01*) and the duration when they waited for the robot (*r=-0.362, n=20, p=0.116*).

The relationship between the NARS index of participants and their comfortability was positive but was not significant (*r=0.071, n=20, p=0.763*).

**Influence of age group and gender**

There was a statistically significant positive correlation between the age and the engagement as measured by gaze duration (*M=3.84, SD=4.34*), *r=.56, n=20, p=0.004*. This is depicted in *Figure 7a*. As age increases, the participants tend to gaze more at the robot during communication.

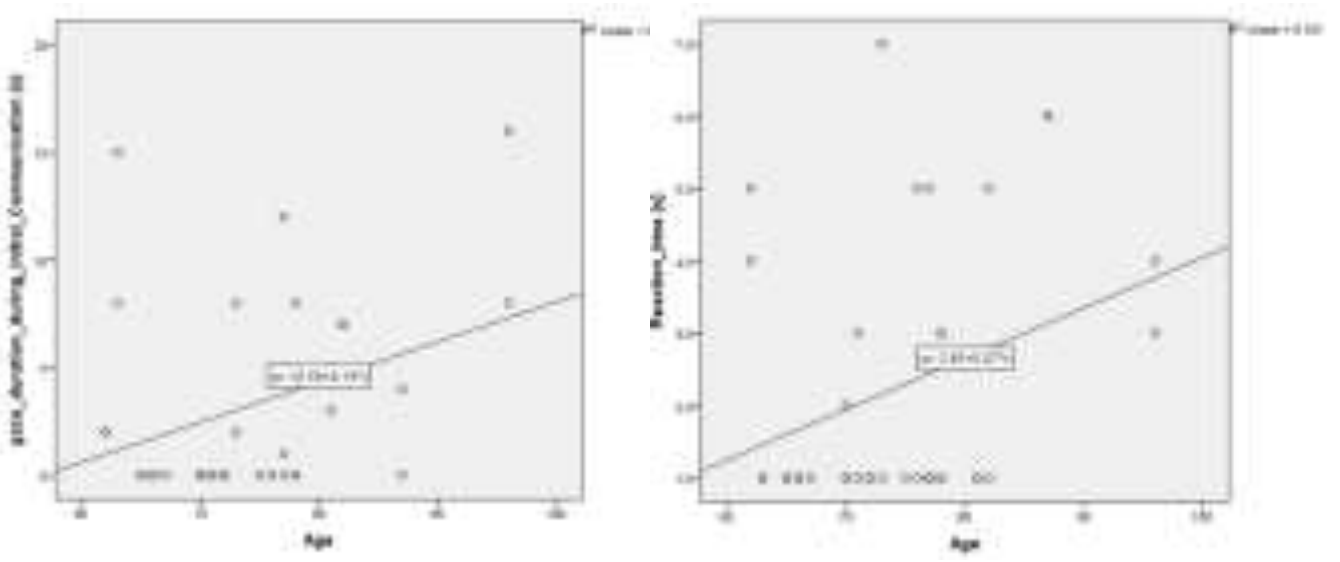

Figure 7: a) Correlation between Age and Gaze duration b) Correlation between Age and Reaction time

There was also a trend between the age and the understanding of the participants as assessed in terms of the number of clarifications made by the participants *(M=0.4, SD=0.5)* and the reaction time in seconds (M=2.27, SD=1.86). The correlation between the age of the participants and the number of clarifications made by the participants was not significant (*r=0.097, n=20, p=0.568*) but there was a fairly significant positive correlation between age and reaction time, (*r=0.32, n=20, p=0.056*). As the age increased, slower response to the robot's instructions were observed. This is presented in *Figure 7b*.

The trend between age and trust, which was measured in terms of the duration when the participants waited for the robot when the robot lost track (*M=2.4, SD=3.79*), was also explored (*Figure 8*a). It was observed that that there is a significant negative correlation between age and the waiting duration (*r=-0.443, n=20, p=0.027)*.

There was no significant correlation between the age of the participants and their level of comfortability as assessed by the number of back glances made to the robot while walking (*M=2.9, SD=3.6, r=-0.287, n=20, p=0.164*) and the total time they spent with the robot (*M=2.9, SD=3.6, r=-0.231, n=20, p=0.267*). Qualitative analysis reveals some negative relationship trend with age (Figure *8b*).

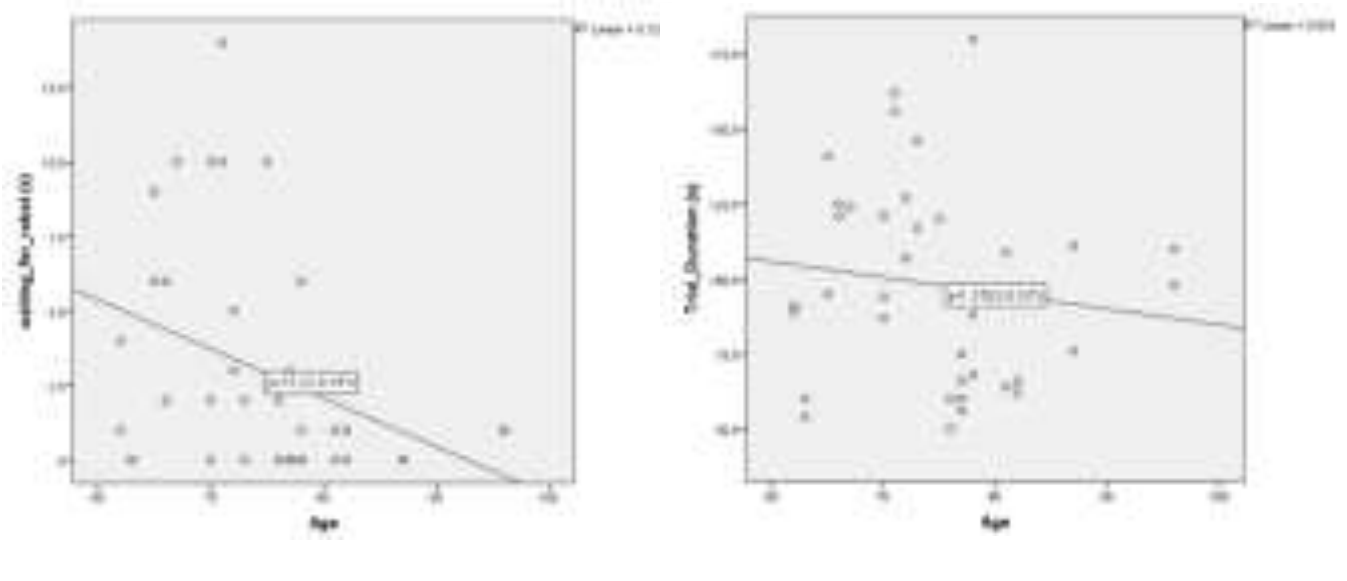

Figure 8: a) Correlation between Age and Waiting time b) Correlation between age and time spent with the robot

With regards to gender analyses, the females (*M=5.43, SE=1.33*) seemed more engaged (*F(1,13)=4.5, p=0.054*) as seen in the gaze duration (*M=3.59, SE=0.005)* compared to the males (*M=1.74, SE=1.47*). The males (*M=1.74, SE=1.47*) also seemed to trust the robot less than the females (*M=1.74, SE=1.47*) as observed in the duration of time spent (*M=2.84, SE=0.837)* with the robot, *(F(1,12)=0.898, p=0.362*).

In terms of understanding, as assessed through the number of clarifications made (*M=0.61, SE=0.13)*, it was observed that the females asked more questions than the males, but this was not statistically significant, *(F(1,23)=0.123, p=0.729*). The differences in the level of comfortability each gender experienced as assessed by the number of times they glanced back at the robot (*M=2.23, SE=0.675*) and the amount of time they spent with the robot (*M=93.24, SE=9.11, p=0.934*) was also not statistically significant (*F(1, 23)=0.007, p=0.934*).

## 8 DESIGN GUIDELINES

This is a first attempt to explore feedback parameters via a series of user studies focusing on a person following robot application for older adults. This sequential user-centered study aimed to ascertain the older adults' user needs and preferences regarding feedback from the robot for this defined task of person following. The implications of this series of sequential studies as relating to improved feedback design are presented in the following subsections.

### 8.1 Transparency Considerations

***Users prefer information on what the robot is doing (LOT 1)***. Hence, ***they do not need the robotic system to be fully transparent***, rather they ***want it to be current and immediate***. They are satisfied with the robot ***communicating just its current actions and status information***. Older adults seem to trust that the robot will know how to handle itself if more information is available or if the state of matters will change despite their initial disposition to the robot as revealed in the NARS index. This agrees with the discussions in [26] where it was hypothesized that the users may prefer less information based on the degree of trust they have developed in the system. Users' preference in our first study also concurs with the design principles for transparency, outlined in [19] where designers were cautioned regarding providing too much information to users. It was emphasized that if such information exceeds the preferences and needs of the users, it may bring frustration and/or confusion.

In order not to limit the participants to receive feedback only for task-related transparency options, participants were asked to suggest additional information that they would want the robot to give. This was to make room for other aspects of transparency relating to the robot such as information on how the robot makes its decisions or principles guiding its actions. The discussion was also intended to address environment-related feedback content such as structure of the environment, constraints in the environment and safety-related information about the environment. Caution was taken to avoid overloading the participants with too many transparency options. Therefore, transparency models connected with teamwork (information on the role of the robot and human in task), human state (information regarding physical, emotional or stress state of the participant) were not mentioned. Participants were asked to point out specific content they would like the robot to give. Participants' responses (*Figure 5*) indicate that they were interested in task-related transparency information (such as 'following', 'stopping'). In addition, some participants wanted the robot to ask about their wellbeing ('greetings', 'how are you?'). These are aspects of the human related model of transparency which participants provided without being directly asked. It supports the significance of 'thinking aloud' sessions recommended in user-centered system designs [27]. The preference for greetings also supports the finding of Sabelli et al [62].

The outcome of the first stage also brings to the fore an interesting contrast in the LOT demands of younger and older adults. In a previous study [33], earlier discussed, where younger adults (aged 18 – 22 years) participated in a user study to examine the effect of transparency and feedback modality on trust, they preferred higher LOTs. This may not only be an age-related trust issue but may also be connected with the embodiment of the robot. The robot in [33] was simulated on a computer desktop and not physically present as used in the current study. This suggests that interacting with a physical robot and observing its performance may have a stronger effect on the users' trust and affect the amount of information (LOT) such user may prefer the robot to provide compared to a simulated robot.

The population in this study may be unique in their LOT demands, but we cannot assure this conclusion since there the studies included a specific task. To establish a stronger mapping between the preferences in this study and that of a wider population of older adults, more extensive studies are recommended as suggested in [63]. These further studies would assess the external validity of this outcome on a larger scale. Studies that examine the possible changes in users' transparency demands such as trust and comfortability adaptation for interacting with a robot occurs over longer periods of interaction.

### 8.2 Feedback Modality Considerations

***Users prefer the robot to communicate with them in voice mode***. The voice, as compared to tone-mode, tends to give the robot a form of personality which enables users to better envision it as an assistant or partner than just a mere machine. This tallied with the findings in [64] where it was highlighted that verbal mode improves perceptions of friendship and social presence. Even though, the outcome in [33] seemed to portray that feedback modality was not significant, the task was different which emphasizes the importance of evaluating the feedback design parameters in specific tasks to ensure applicability to such tasks as recommended in the study [33] and in [26]. This also agrees with the recommendations in [19], regarding designing communicative interfaces to ensure that the feedback modality fits the needs and preference of the user in defined tasks. In the current task, where the robot follows the user, the feedback modality (voice feedback) tends to keep the users more engaged with the robot which is one of the variables that indicates a potential of improvement in the quality of interaction.

Identifying a primary means of feedback was crucial in this study particularly in connection with the preferred content of the feedback. Providing multiple modalities can be explored as a next stage, with the possibility of including haptic feedback. However, considerations must be made regarding the cognitive peculiarities of the older adult users which influence the number of sources of information they can process per time [11], [58]. It is pertinent that the older adults are not overloaded with information. There is the potential of adaptable modality selection [37], [40], [65] which may provide the option of user-defined modality preferences based on the complexity of task, human physical or cognitive state, performance and environmental related factors. This would give the older adult allowance to further personalize their feedback modality preferences which aligns with the goal of meeting the needs, preferences, capabilities and limitations of users [40], [66], [67].

### 8.3 Feedback Timing Considerations

***Continuous feedback, at short intervals, was preferred*** by the participants. It seemed to provide them with better awareness regarding the state of the interaction compared to discrete feedback used in previous studies. This was in conformity with previous studies where continuous feedback timing was found to improve users' awareness [39], [68] even though these studies were not focused on older adult users. The outcome of this stage therefore highlights a crucial feedback design component of ***proving minimal information*** (*LOT1*) ***continuously at short intervals***. We would however recommend that this preference be treated with caution, as preference of the users could change with the complexity of the task or the duration of interaction. On a long term basis, participants may adapt their level of trust in the robot and this may make them rely on the robot more, such that longer intervals between continuous feedback messages may be preferred.

Different degrees of involvement of the robot in the task may also have an influence on the frequency of information required by the participants [19], [25]. Users may require information less frequently from a robot that is more autonomous than one which is more dependent on the user for each action. This concept of the influence of the robot's level of autonomy on feedback timing in the context of person-following task for older adults requires further exploration.

### 8.4 Predisposition Considerations

The results observed from the correlation analyses of the effect of the initial attitude of participants as indicated by the NARS responses revealed the impact that the predisposition of the participants towards robots could have on their interaction with the robot. This reflects previous findings [55] where it was explained that the initial attitude of the participants affected the manner they evaluated the robot which then influences the interaction [55]. Bishop et al [69] had highlighted the negative influence that subjective negative affect could potentially have in the interaction with a robot. This could be responsible for the trends seen where participants who had a more negative initial attitude towards the robot even before interaction (with or without the feedback) seemed to understand and trust it less.

It is therefore important to include some form of introductory session by the robot to better prepare the older adult for the interaction. The feedback design parameters and interfaces should make allowance for such user-friendly initial introductions before the actual task implementation with the robot. The older adults should however, also be given the option to skip this session if they are already familiar with the robot so that this introduction session does not induce boredom in the interaction.

### 8.5 Gender and age considerations

Results revealed that age and gender could influence the perception, preferences and attitudes of the participants towards the robot. Even though, there may be some intra-individual differences that may also be responsible for some of the observations made [46], results in this study reveal that the inter-individual differences stemming from age and gender are worth considering in the design. Thus, in the process of developing a user-centered feedback design, the preferences of women should be considered differently from that of the men. The feedback parameters should be tuned to suit the preferences of older adult users in different age categories.

For example, with regards to engagement, the trend reveals that the older old adults tend to be more engaged with the robot compared to their younger counterparts as seen in the gaze duration analyses. This could be connected with novelty effect where a larger percentage of the younger old adults may have been more familiar with some form of related technologies compared to the older old adults[8]. This could inspire more attraction to the robot, and thus engage them more. This agrees with previous findings [69] where it was shown that familiarity with related technology negatively correlates with the attitude and intention to interact with the robot. Those who were more familiar with related technology may find the interaction less enjoyable and thus may not be as engaged as those who are less familiar [69]. Attention has to be placed on measures to improve the engagement in the younger older adult category.

It was also understandable that the older-old adults had a longer reaction time when interacting with the robot as seen in the correlation of age with understanding. Several older old adults do not have as much experience with technology as the younger old adults as observed in Heerink's study [70]. It was additionally established in the study that experience with related technology aids the use of a system [70]. This could explain the reason why the younger old adults who likely had more experience with technology seemed to have a better understanding of the robot's operation compared to the older adults. It also emphasizes the importance of adapting some of the feedback parameters such as clarity, repetition of instruction, rate of feedback to aid the understanding of the older old adults.

The older old adults seemed to trust the robot more than the younger old adults as seen in their waiting time. This also agrees with the discussions in [69] where it was stated that younger users who may be more familiar with related technology were more aware of the robot's limitations and therefore may have felt less safe around the robot. This could be affect the trust the more familiar people felt around the robot. Even though Broadbent et al [71] mentioned that some older people may show more negative emotions towards robots, this study reveals that familiarity may not necessarily improve the trust index. However, if the robot constantly informs the user on its capabilities, this could have some influence on the users' willingness to trust the capabilities of the robot as observed in [70]. It also brings an important consideration regarding feedback design to the fore: informing users of the capabilities of the robot and demonstrating such capabilities. This form of communicative attribute coupled with reliable performance of the robot as emphasized by Hancock et al [72] in their study of factors influencing trust in HRI, can potentially help the users trust the robot better. It may also improve the comfortability trend at all age categories as established in the Almere model [73] showing perceptual influences on acceptance of a robot by older adults.

Gender had also been found in previous studies to have a significant influence on the interaction with the robot [69], [70], [74]. This was confirmed in the current study where the females were seen to be more engaged to the robot than the males and also seemed to ask more questions to clarify their understanding of the robot better. The females also seemed to trust the robot more as seen in the time they waited for the robot. They seem to trust that the robot would perform correctly even when it delayed or lost track. Even though, Heerink [74] and de Graff [73] associated anxiety with the females' interaction with a robot, the current study agrees with earlier findings by Shibata et al [75] where it was stated that females are more comfortable around robots. This could potentially influence trust positively. Several reasons could be responsible for this disparity which includes context and type of robot. However, the reason cannot be fully established from this study due to the limited sample. But it highlights the need to further explore the expectations and needs of the different genders such that the feedback design could be tuned to meet possible gender preferences.

Gender and age category of the older adult users should therefore be adequately considered in order to meet the specific needs in the different groups that make up the older adults' population.

### 8.6 Design Implications, Limitations and Future Work

While evaluating the effectiveness of the feedback design, we observed that the users were more engaged with the robot, understood the robot more, and better trusted the robot when it communicated with them using the implemented feedback compared to situations where the robot followed without such feedback. This confirms hypotheses *H1 – H3*. The responses of the participants in the questionnaires also revealed that they were more comfortable communicating with the robot when the feedback was implemented. This confirms hypothesis *H4*. The feedback was designed to match the perceptual demands of the target users. The outcome supports the proposition in literature that such ***user-centered feedback design can increase the quality of interaction***.

One of the limitations of this study is that the feedback design was evaluated for a single task scenario. The feedback was not evaluated in multiple task situations with varying environmental variables such as noise and space type. Evaluation of the feedback design parameters is also recommended for an extended period of time in order to assess the preferences of the older adults as the novelty effect wears off. It is also recommended that training be conducted for older adult users as naïve users, regarding how the feedback interfaces operate in order to maximize the interaction quality. These are crucial factors that should be considered in future work to improve the robustness of the feedback design. The outcome of this study provides some guidelines and recommendations that could be useful while conducting more extensive studies on feedback design guidelines in person following robots that will accommodate user needs in eldercare.

## ACKNOWLEDGEMENTS

This research was supported by the EU funded Innovative Training Network (ITN) in the Marie Skłodowska-Curie People Programme (Horizon2020): SOCRATES (Social Cognitive Robotics in a European Society training research network), grant agreement number 721619 and by the Ministry of Science Fund, grant agreement number 47897. Partial support was provided by Ben-Gurion University of the Negev through the Helmsley Charitable Trust, the Agricultural, Biological and Cognitive Robotics Initiative, the Marcus Endowment Fund, the Center for Digital Innovation research fund, the Rabbi W. Gunther Plaut Chair in Manufacturing Engineering and the George Shrut Chair in Human Performance Management.